\documentclass[11pt]{article}

\usepackage[final]{acl}
\usepackage{amsmath}
\usepackage{graphicx}
\usepackage{caption} 
\usepackage{times}
\usepackage{latexsym}
\usepackage{subcaption}
\usepackage{booktabs}
\usepackage{multirow}
\usepackage{pifont}
\usepackage{subcaption}
\usepackage[T1]{fontenc}

\usepackage[utf8]{inputenc}

\usepackage{microtype}

\usepackage{inconsolata}

\usepackage{graphicx}
\usepackage[most]{tcolorbox}
\usepackage{stfloats}   
\tcbset{boxsep=0pt}

\newtcolorbox{wideheaderbox}[2][]{%
  enhanced,
  colback=gray!3, colframe=black, boxrule=0.6pt, arc=2mm,
  left=3mm, right=3mm, bottom=2.5mm,
  top=8mm,                    
  width=\textwidth,
  overlay unbroken={%
    \fill[black,rounded corners=2mm]
      (frame.north west) rectangle ([yshift=-7mm]frame.north east);
    \node[anchor=west]
      at ([xshift=4mm,yshift=-3.5mm]frame.north west)
      {\bfseries\color{white}\strut #2};      
  },
  overlay first={%
    \fill[black,rounded corners=2mm]
      (frame.north west) rectangle ([yshift=-7mm]frame.north east);
    \node[anchor=west]
      at ([xshift=4mm,yshift=-3.5mm]frame.north west)
      {\bfseries\color{white}\strut #2};
  },
  #1
}

\title{Exploring Generative Process Reward Modeling for Semi-Structured Data: A Case Study of Table Question Answering}

\author{Lei Tang$^{2}$ \hspace{3mm} Wei Zhou$^{1,3}$ \hspace{3mm}
    Mohsen Mesgar$^{3}$  \\
$^1$University of Augsburg, Germany \hspace{5mm} \\
$^2$Institut f\"{u}r Maschinelle Sprachverarbeitung, University of Stuttgart, Germany \\ 
$^3$Bosch Center for Artificial Intelligence, Renningen, Germany \\ 
\texttt{letang9-c@my.cityu.edu.hk} \hspace{3mm} 
\texttt{\{wei.zhou3|mohsen.mesgar\}@de.bosch.com}\\ %
}

\begin{document}
\maketitle
\begin{abstract}
Process reward models (PRMs) enhance complex reasoning in large language models (LLMs) by evaluating candidate solutions step-by-step and selecting answers based on aggregated step scores. While effective in domains such as mathematics, their applicability to tasks involving semi-structured data, like table question answering (TQA), remains unexplored. TQA poses unique challenges for PRMs, including abundant irrelevant information, loosely connected reasoning steps, and domain-specific reasoning. This work presents the first systematic study of PRMs for TQA. We evaluate state-of-the-art generative PRMs on TQA from both answer and step perspectives. Results show that PRMs that combine textual and code verification can aid solution selection but struggle to generalize to out-of-domain data. Analysis reveals a weak correlation between performance in step-level verification and answer accuracy, possibly stemming from weak step dependencies and loose causal links. Our findings highlight limitations of current PRMs on TQA and offer valuable insights for building more robust, process-aware verifiers.
\end{abstract}
\section{Introduction}
Recent advances in process reward models (PRMs) have demonstrated remarkable improvements in the reasoning capabilities of large language models (LLMs) \cite{wang-etal-2024-math, yin-etal-2025-dynamic, Ong2025TrainingVP}. 
PRMs enable sampling-based test time scaling (TTS) by grading multiple solutions step-wise. The best solution is then selected based on aggregated step scores, e.g., average step scores.
This verification-driven paradigm has proven highly effective in domains such as math \cite{zhao2025genprmscalingtesttimecompute, zhang2025lessonsdevelopingprocessreward,zhang-etal-2025-process}, visual \cite{Ong2025TrainingVP,Wang2025VisualPRMAE}, and textual QA \cite{yin-etal-2025-dynamic}.
However, to the best of our knowledge, no previous work explores the effectiveness of PRMs on tasks using semi-structured data, e.g., tables. 
This raises a natural and important question: \textit{Is the success of PRMs in domains like math transferable to other complex reasoning tasks, where semi-structured data are involved?}
To answer the question, we focus on table question answering (TQA), where a system needs to solve a question based on information in a table. 

\begin{figure}
    \centering
    \includegraphics[width=1\columnwidth]{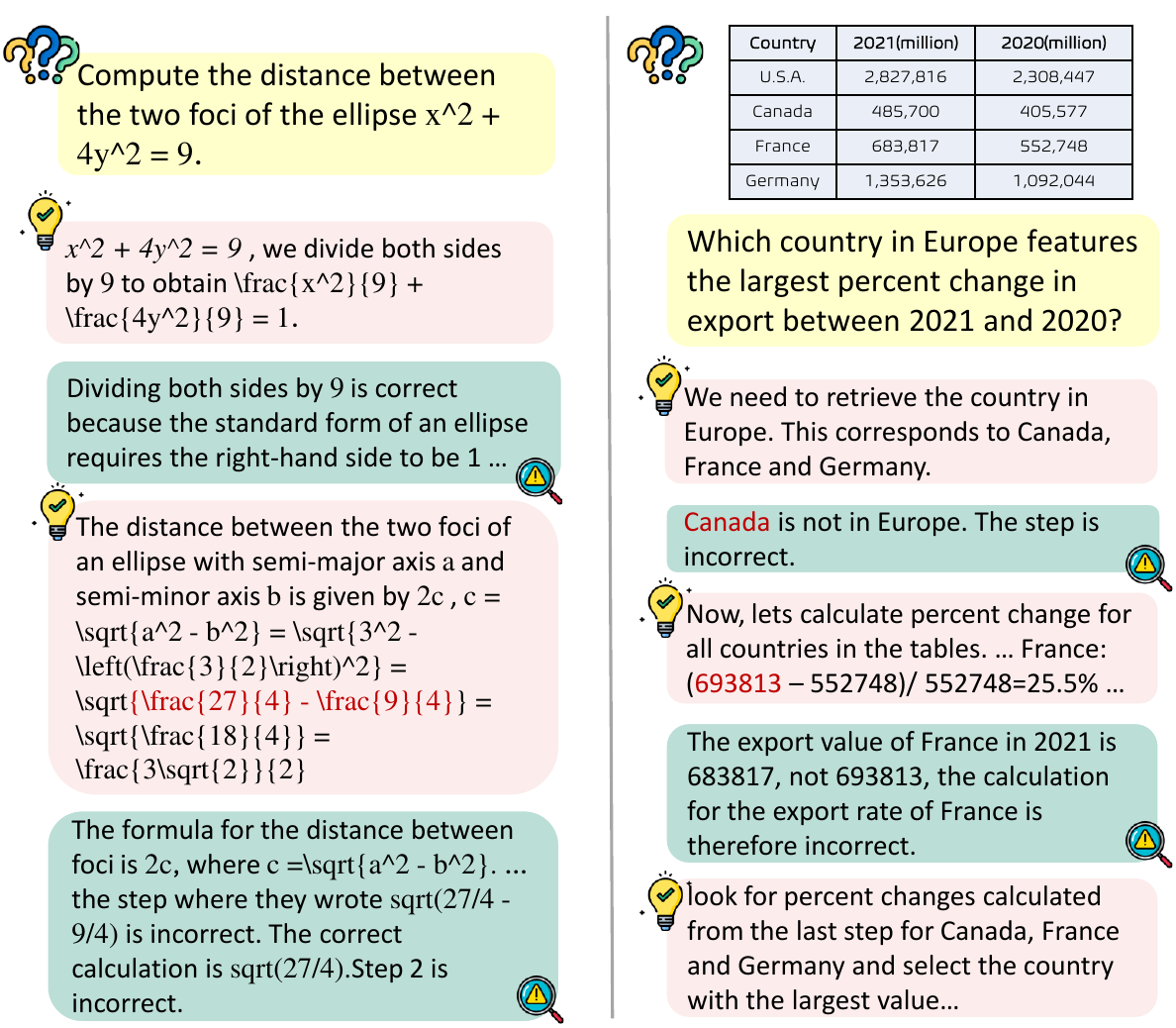}
    \caption{Comparing the verification process for a math (left) and a TQA problem (right). Examples draw from \citet{zhao2025genprmscalingtesttimecompute} and \citet{zhou2025efficientmultiagentcollaborationtool}. Compared to math, not all information in TQA is relevant. }
    \label{fig:overview}
\end{figure}
TQA differs fundamentally from math problems. In math, most given information is relevant, and reasoning steps are tightly linked in deductive chains. 
In contrast, tables in TQA can be large with abundant irrelevant information for solving a question. 
Moreover, the reasoning steps are often loosely connected, i.e., later steps may not strictly follow from earlier ones and can start a new topic. Consider ``Step 1'' and ``Step 2'' on the right side of Figure \ref{fig:overview}. 
Nevertheless, both tasks require multi-step and complex reasoning to be solved. 

We hypothesize that TQA can benefit from PRM-based verification. Unlike existing TQA approaches, which predominantly optimize policy models through prompt engineering \cite{zhou2025efficientmultiagentcollaborationtool, wang2024chainoftableevolvingtablesreasoning}, or fine-tuning \cite{zhang-etal-2024-tablellama,deng-mihalcea-2025-rethinking}, PRMs are model agnostic: they can be applied to outputs from any policy models and rewards given by PRM can be used to further improve performance of policy model via reinforcement learning with verifiable rewards (RLVR) \cite{Su2025CrossingTR}.  
Moreover, solutions selected by PRMs reflect both answer and step quality, compared to majority voting that considers only answer frequencies \citep{liu-etal-2024-rethinking,zhou2025efficientmultiagentcollaborationtool}. Note that the effectiveness of majority voting depends strongly on the base model's capability: it has to produce the correct answer in most sampled outputs. By contrast, PRMs are specifically trained for step-level evaluation. Last but not least, PRMs can enhance system trustworthiness, providing human-readable rationales for verification decisions on step correctness.

Yet, applying PRMs to TQA is non-trivial. Retrieval noise, loosely connected reasoning steps, and domain-specific reasoning patterns may limit their effectiveness. 
As an initial step, this work assesses the potential of applying PRMs for step-level verification in TQA. We explore to what extent such a step-based evaluation paradigm assists in solution selection, with a focus on PRMs that generate human-understandable rationales. Specifically,
we evaluate multiple PRM variants from both answer-level and step-level perspectives against strong baselines, including self-consistency \cite{Wang2022SelfConsistencyIC} and LLM-as-a-judge \cite{Gu2024ASO}. Our results show that PRMs can aid in selecting high-quality solutions for TQA, particularly when combining textual and code verification. However, their effectiveness remains limited, with notable weaknesses such as poor generalization to out-of-domain datasets.

Through qualitative and quantitative analysis of verification chains, we uncover a weak correlation between step verification and answer correctness: better average step scores do not necessarily yield higher answer accuracy in TQA. This stems from characteristics of the problem profile, such as weak step dependence and loose causal relationships between reasoning steps. By bridging PRMs with TQA, our work clarifies when and why PRMs succeed or fail, and provides insights for designing more robust, process-aware systems for complex multi-step reasoning tasks.
Code is available.\footnote{\url{https://github.com/boschresearch/TQA-PRM}}

\section{Examining PRMs on TQA.}
This section describes the PRMs we examine, the training setup, baselines, TQA datasets for training and evaluation, as well as evaluation protocols.  
\paragraph{Problem Formulation.} 
Let $T$ denote a table, $Q$ a natural language question, and $S$ a set of candidate reasoning paths generated by a policy model $M_p$, where each path produces a predicted answer $A'$. 
A PRM evaluates each reasoning step $s_i \in S$ by generating a natural language rationale $v_i \in V$ and assigning a scalar reward $r_i \in R$. 
The goal is to identify the reasoning path $S^\ast \in S$ that maximizes the likelihood of producing a correct answer, i.e., $A' = A$, where $A$ is the ground-truth answer.

  

\paragraph{Examined PRMs.}
We focus on state-of-the-art generative PRMs that output both step-level rewards and rationales explaining these rewards. \textsc{GenPRM} \cite{zhao2025genprmscalingtesttimecompute} verifies each step through a two-stage process: (1) a textual verification stage, in which the PRM examines a step using chain-of-thought (CoT) reasoning  \cite{wei2023chainofthoughtpromptingelicitsreasoning}, and (2) a  code verification stage, where Python code is generated and executed for the current step. Step correctness is determined based on the CoT reasoning, generated code, and its execution results. \textsc{Textual PRM} \cite{zhao2025genprmscalingtesttimecompute} is an ablation of \textsc{GenPRM} that removes the code verification stage, relying solely on CoT analysis. \textsc{Multi-Dimensional Analysis (MDA)} \cite{she2025rprmreasoningdrivenprocessreward} is a textual verifier that evaluates each step from five distinct perspectives: restatement, data check, logical consistency, numeric accuracy, and format. Finally, \textsc{Rephrase–React–Analysis (RRA)} \cite{deng2023rephrase} rephrases the current step as a question, solves the question using the ReAct method \cite{Yao2022ReActSR}, and then provides a step-by-step analysis. Implementation details are provided in Appendix~\ref{appendix:implementation}.

\begin{figure*}[t]
  \centering
   \begin{subfigure}[b]{0.49\textwidth}
    \centering
    \includegraphics[width=\linewidth]{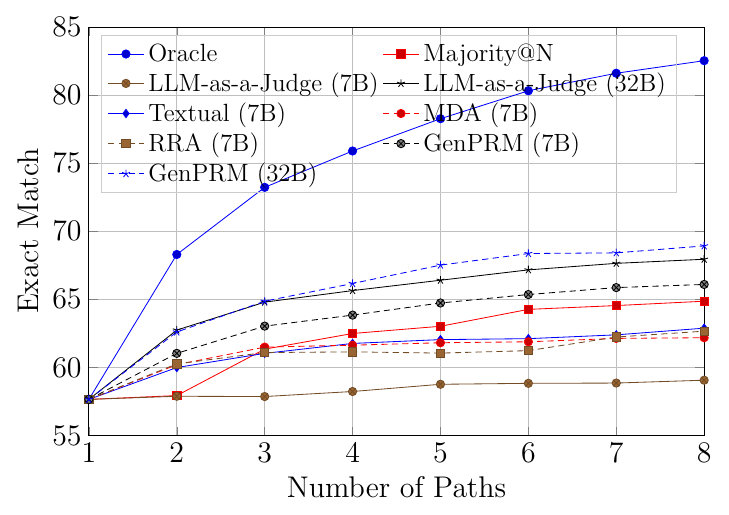}
    \label{perfor:wtq}
     \end{subfigure}
     \hfill
   \begin{subfigure}[b]{0.49\textwidth}
    \centering
    \includegraphics[width=\linewidth]{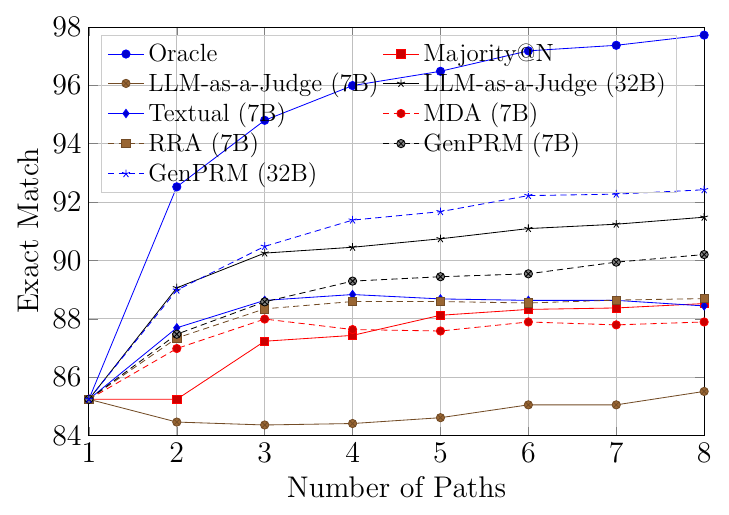}
    \label{perfor:tabfact}
   \end{subfigure}
   \caption{EM measured by \textit{Best@N} across varying numbers of paths. Left: WTQ. Right: Tabfact. \textsc{Oracle} always selects correct answers if they exist. \textsc{Majority} selects the most frequent predictions as answers.  }
   \label{fig:main_results}
\end{figure*}

\paragraph{Training PRMs for TQA.}
We use the training sets of WTQ \citep{pasupat-liang-2015-compositional} and TabFact \citep{chen2020tabfactlargescaledatasettablebased} as our source problem sets for obtaining reasoning paths. 
For each question table pair, we sample $N = 8$ reasoning paths from Qwen2.5-7B-Instruct \citep{qwen2025qwen25technicalreport} with a temperature of 0.6.
For verification generation with \textsc{GenPRM}, we adopt the default hyper-parameters from \citet{zhao2025genprmscalingtesttimecompute} and perform relative progress estimation (RPE), defined as the ratio between the scores of $s_i$ and $s_{i-1}$. These scores are approximated using Monte Carlo sampling \cite{wang2024mathshepherdverifyreinforcellms}. We employ QwQ-32B \cite{qwq32b} as the rationale generation model to produce both textual analyses and Python code. The normalized probability of the \texttt{Yes} token is taken as the final step reward.
If the rewards generated by RPE and the rationale model disagree, the instance is discarded. 
We also filter out incomplete outputs, instances with improper formatting, and cases where the predicted label is neither \texttt{Yes} nor \texttt{No}.
To generate training data for textual verification, QwQ-32B is used. We provide analysis of generated data quality in Appendix~\ref{appendix:quality_analysis}.
For PRM training, we fully fine-tune DeepSeek-R1-Distill-Qwen-7B \cite{DeepSeekAI2025DeepSeekR1IR}, chosen for its large context length, on the filtered dataset. The training data are formatted as standard multi-turn conversations. Training parameters can be found in Appendix \ref{training_para}.

\paragraph{Baselines and Evaluation.}    
We compare against the following baselines:    
\textbf{\textsc{Oracle}} which selects the correct answer from the candidate set, serving as an upper bound on verification performance.  
\textbf{\textsc{Majority@N}} chooses the most frequent answer among $N$ candidates; in the case of a tie, the first prediction is selected.  \textbf{\textsc{LLM-as-Judge}} uses the QwQ-32B model as an LLM judge, with the prompt provided in Appendix~\ref{prompts}.  
We adopt two evaluation paradigms: \textbf{answer-focused} and \textbf{process-focused}.  
In the \textit{answer-focused} setting, we apply a \textit{best-of-$N$} strategy: each sampled reasoning path is scored by verifiers, and the final answer is taken from the path with the highest average step score.    
Performance is measured using exact match (EM).  
We report results in both \textbf{in-domain} and \textbf{out-of-domain} scenarios, evaluating on the test sets of WTQ~\citep{pasupat-liang-2015-compositional}, TabFact~\citep{chen2020tabfactlargescaledatasettablebased}, CRT~\citep{zhang-etal-2023-crt}, and SciTab ~\citep{lu-etal-2023-scitab}. Dataset statistics are shown in Appendix \ref{appendix:data_license}.
In the \textit{process-focused} settings, we manually annotate 300 instances which are randomly sampled from the four datasets, resulting in 1916 steps (1366 correct and 550 incorrect). 

\section{Results and Analysis}

Figure~\ref{fig:main_results} presents EM of various PRMs and baseline approaches across different path sampling budgets on WTQ and TabFact. Results for step-level verification are shown in Table \ref{tab:process_evaluation}.

\paragraph{\textsc{GenPRMs} achieve promising performance compared to other methods, while falling short of \textsc{oracle} for in-domain settings.}
From Figure \ref{fig:main_results}, we observe that \textsc{GenPRMs} consistently delivers the strongest results \textbf{compared with examined generative PRMs}.
We attribute it to its integrated code verification component. A detailed analysis is provided in the following paragraph. 
\textbf{In comparison to baseline methods}, both the 7B and 32B variants of \textsc{GenPRMs} substantially outperform majority voting, underscoring the benefit of process-level verification for answer selection. 
Nevertheless, the fine-tuned 7B \textsc{GenPRMs} lags behind the 32B LLM-as-a-judge. Though scaling \textsc{GenPRMs} to 32B  closes this gap, performance advantages over LLM-as-a-judge remain limited. 
Across all methods, a substantial performance gap remains compared to \textsc{oracle}, indicating considerable room for improvement in leveraging multiple candidate solutions. We discuss the efficiency of different methods in Appendix \ref{efficiency}.

\paragraph{Incorporating code verifier improves PRM performance in both answer and step correctness.}
From the answer-level perspective, a comparison between \textsc{CoT PRM} and \textsc{GenPRM} in Figure~\ref{fig:main_results} shows that \textsc{GenPRM} substantially outperforms \textsc{CoT PRM}. Since \textsc{CoT PRM} is trained on the same data as \textsc{GenPRM} but without the code verification component, this performance gain can be attributed to the incorporation of code verification. 
From the process-level perspective, we examine step-level correctness by comparing rewards derived from pure textual analysis with those obtained after applying code verification in \textsc{GenPRM}. The results, presented in the last four rows of Table~\ref{tab:process_evaluation}, demonstrate that incorporating code verification markedly improves step-level correctness verification across datasets and model sizes. 

\begin{table}[t]
\small
\setlength{\tabcolsep}{3pt}
    \centering    
\begin{tabular}{lcccc}
 \toprule
\textbf{Method} & WTQ & TabFact &CRT &SciTab  \\
\midrule
LLM-as-a-Judge (32B) &90.29 &88.56 &81.60 &84.82\\
Textual PRM (7B) & 81.06 & 80.35 &71.58 &74.70 \\
\textsc{MDA} (7B) &80.79 &81.25 &72.10&72.61\\
\textsc{RRA} (7B) &71.52 &84.82 &70.37 &77.08\\
GenPRM w/o code (7B) &80.99&80.29&71.58&75.00\\
GenPRM (7B) & 91.38 & 90.13 &77.58 &84.93  \\
GenPRM w/o code (32B) &84.79 &87.50 &79.80 &80.92 \\
GenPRM (32B) &\textbf{93.71} &\textbf{94.72} & \textbf{83.22}& \textbf{85.71}\\
\bottomrule
\end{tabular}
\caption{Process verification accuracy of PRMs and LLM-as-a-Judge on four TQA datasets.\textsc{RRA} stands for Rephrase, React, Analysis (7B). \textsc{MDA} stands for multi-dimensional analysis.}
\label{tab:process_evaluation}
\end{table}

\paragraph{When Code Helps and Hurts.}
We conduct a manual analysis to characterize when code-based verification is beneficial and when it is not. Specifically, we review 60 randomly sampled steps on which GenPRM with code and without code disagree (e.g., a step labeled 1 by GenPRM with code but 0 by the model without code). Among these 60 instances, GenPRM with code assigns the correct label in 43 cases and an incorrect label in 17 cases.
Within the correctly classified instances, we observe two recurring patterns in which code execution provides substantial benefits: (1) \textbf{Numerical operations.} Executing code supports precise and quantitative reasoning (e.g., counting, aggregation, and threshold checks), thereby reducing ambiguity and limiting error propagation. (2) \textbf{Targeted retrieval.} Programmatic access to specific rows or columns provides accurate relevant evidence and enables  verification of individual cell values.
We also identify two common failure modes of code-based verification: (1) \textbf{Exhaustive retrieval.} The program retrieves large portions of a table (or even a full table) even when the claim depends on a small subset, implicitly shifting the burden to long-context recall rather than selective filtering. (2) \textbf{Index misunderstandings.}  The model sometimes confuses row indexing conventions, inconsistently switching between absolute row indices (including the header) and row indices that exclude the header, which leads to incorrect cell retrieval.

\paragraph{Fine-tuned PRMs with code verification are not robust on out-of-domain datasets.}
As shown in Figure~\ref{fig:main_results}, \textsc{GenPRM} consistently outperforms majority voting on in-domain datasets. However, this advantage diminishes in out-of-domain settings, with performance dropping to 53 vs.56 on CRT and 60 vs.62 on SciTab, as reported in Table~\ref{tab:full_results}. We attribute this gap to differences in problem complexity: CRT contains highly complex questions, while SciTab requires specialized scientific knowledge. These challenging scenarios often lead the policy model to generate reasoning paths containing spurious or noisy steps, which hinder effective verification. An analysis of variance in reasoning complexity across datasets is provided in Appendix~\ref{appendix:variance_dataset}.
In summary, supervised fine-tuned PRMs are prone to generalization issues. Mitigating this limitation may require more balanced fine-tuning data that reflects varying verification difficulties, or additional post-training techniques to enhance the generalizability of PRMs.

\paragraph{Process-level evaluation highlights the competitiveness of \textsc{GenPRM}.}
While the fine-tuned 7B \textsc{GenPRM} under-performs the 32B LLM-as-a-judge in answer evaluation, process-level verification reveals a different trend: 7B \textsc{GenPRM} attains comparable or higher accuracy across most datasets, with the exception of CRT (Table~\ref{tab:process_evaluation}). This suggests that strong step-level verification does not necessarily translate into improved answer accuracy. We elaborate on this point in the next paragraph. Moreover, purely text-based PRMs consistently lag behind \textsc{GenPRM}, underscoring the benefit of integrating code verification.

\paragraph{Weak correlation between PRM verification and answer correctness in TQA.}
\begin{figure}[t]
  \centering
  \begin{subfigure}[t]{0.24\textwidth}
    \centering
    \includegraphics[width=\linewidth]{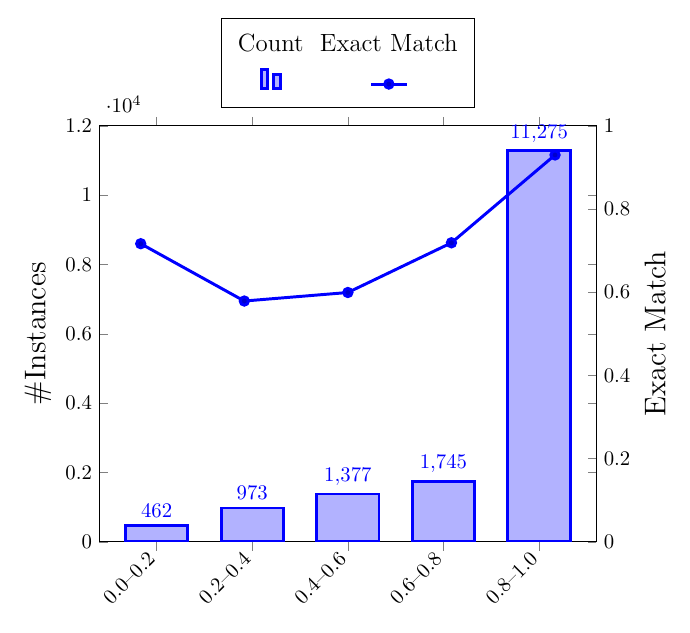}
    \caption{GenPRM (7B) on Tabfact}\label{fig:cor_tabfact}
  \end{subfigure}\hfill
  \begin{subfigure}[t]{0.24\textwidth}
    \centering
    \includegraphics[width=\linewidth]{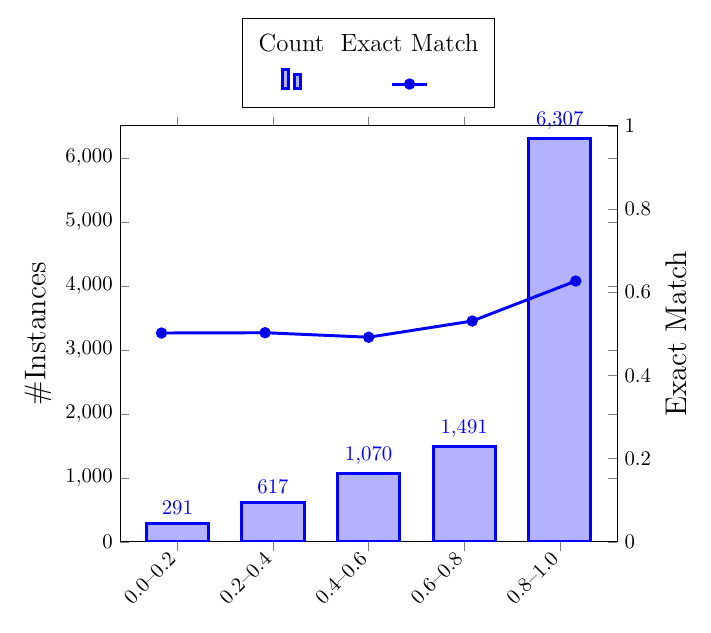}
    \caption{GenPRM (7B) on SciTab}\label{fig:cor_scitab}
  \end{subfigure}
  \caption{Counts per confidence bin (bars, left axis) and Exact Match (line, right axis). Bins are PRM score intervals: $[0,0.2),[0.2,0.4),[0.4,0.6),[0.6,0.8),[0.8,1.0]$. It shows that average step scores are not strongly correlated with answer correctness.}
  \label{fig:correlation}
\end{figure}
Our manual analysis reveals that average step scores are not strongly correlated with answer correctness. Figure~\ref{fig:correlation} confirms this trend by plotting EM accuracy across bins of average step scores. We observe that solutions with higher average step scores do not necessarily feature higher EM accuracy. 
A likely reason is that averaging step-level rewards treats all steps as equally informative, regardless of their causal contribution to the final answer. Consequently, (i) chains containing many irrelevant but locally correct steps receive inflated scores, whereas (ii) chains with a local error that does not affect the final answer are over-penalized. Examples of both cases are provided in Appendix~\ref{case_study}. 
Unlike in math, reasoning chains in TQA are more loosely linked, so an error at step $i$ does not necessarily propagate to subsequent steps or the final prediction. This reduces the effectiveness of step verification for answer selection.
Nevertheless, PRMs remain valuable; the core bottleneck lies in the current answer-focused evaluation paradigm and the absence of process-annotated evaluation datasets for TQA. We advocate future work on process-level evaluation and modeling approaches that strengthen step dependencies and assess each step by both its local correctness and its causal contribution to the final answer, for example, by filtering irrelevant steps.
\paragraph{Potential Alternative Aggregation Strategies.} We observe that simple uniform averaging is suboptimal, as it implicitly assumes that all reasoning steps contribute equally to the final answer; this assumption allows irrelevant or non-causal steps to influence the estimated solution quality and obscure the impact of steps that are truly determinative. Beyond uniform averaging, we explore alternative aggregation strategies that more accurately reflect answer quality in TQA. A simple and intuitive approach is to assign greater weight to the final reasoning steps. When evaluated using a 32B coding verifier, this strategy yields consistent improvements of +3.56\% EM on WTQ, +1.14\% on TabFact, +4.55\% EM on CRT, and +2.69\% on SciTab. In addition, relevance- and causality-aware aggregation strategies represent promising directions for future study. A relevance-aware strategy assigns a relevance score to each step, indicating the step’s relevance to solving the problem. Similarly, a causality-aware strategy assigns a causality score to each step, indicating whether the step influences the final outcome. Incorporating these weights when computing a solution score can down-weight steps that do not contribute to the final answer, thereby leading to a score that more faithfully reflects answer quality.

\section{Conclusions}
In this paper, we examined four generative PRMs on TQA and compared them to strong baselines such as LLM-as-a-judge and majority vote from both answer and process evaluation perspectives. We found that 
PRMs that combine textual and code verification can aid solution selection but struggle to generalize to out-of-domain data. 
Our analysis suggests that the primary bottleneck is not step-level verification, but rather the aggregation of step rewards. Uniform averaging is inadequate for TQA as it assigns equal weight to all steps, regardless of their relevance to the problem or their causal influence on the final answer. Future work can therefore investigate relevance-aware and causality-aware aggregation strategies to generate solution scores that more faithfully reflect answer quality.
\section*{Limitations}
To examine the effectiveness of generative PRMs over semi-structured data, we choose table question answering as a proxy task for initial exploration. Nevertheless, there exist other tasks involving semi-structured data, such as table generation or table summarization. Future studies can explore these tasks to validate the effectiveness of PRMs. Moreover,  we mainly train PRM using the Qwen models, following the practice of PRM training in previous work \cite{zhao2025genprmscalingtesttimecompute}. Though we experiment with varying model sizes, results from a different model family might be insightful.

\bibliography{custom}

\appendix
\section{Appendix}
\subsection{Implementation and Adaptation Details}
\label{appendix:implementation}
We adapt five perspectives from \citealp{she2025rprmreasoningdrivenprocessreward}—previous steps analysis, now step analysis, data source analysis, consistency analysis, and calculation analysis—into ours: restatement, data check, logical consistency, numeric accuracy, and format, as shown in Figure~\ref{fig:multi-dimensional}. For RRA, we follow \citealp{deng2023rephrase}: the PRM first rephrases the current step, then solves the rephrased question using the ReAct paradigm \cite{Yao2022ReActSR}, and finally provides a step-by-step analysis conditioned on the current step, the rephrased problem, and the ReAct trace, as shown in Figure~\ref{fig:rephrase}.

\subsection{Data Quality Analysis}
\label{appendix:quality_analysis}
\begin{table}[ht]
    \small
    \setlength{\tabcolsep}{6pt}
    \centering    
    \begin{tabular}{lcccc}
    \toprule
    \multirow{2}{*}{Label} 
    & \multicolumn{2}{c}{After Filtering} 
    & \multicolumn{2}{c}{Before Filtering} \\
    \cmidrule(lr){2-3} \cmidrule(lr){4-5}
    & \#Steps & \% & \#Steps & \% \\
    \midrule
    Correct   & 254 & 94.42 & 240 & 84.21 \\
    Incorrect & 15  & 5.58  & 45  & 15.79 \\
    \bottomrule
    \end{tabular}
    \caption{Manual evaluation of reasoning-step correctness before and after the GenPRM filtering process. Filtering substantially improves data quality, increasing step-level accuracy from 84.21\% to 94.42\%.}
    \label{tab:data_quality}
\end{table}
Training data quality may partially account for the limited performance of GenPRM on TQA. To examine this, we randomly sample 50 reasoning paths before and after the GenPRM filtering procedure and manually assess their correctness.
As shown in Table~\ref{tab:data_quality}, the filtering process effectively removes incorrect steps. After filtering, which produces the data used for PRM training, the step-level accuracy reaches 94\%, indicating overall high data quality. This suggests that data noise is unlikely to be the primary bottleneck for GenPRM performance on TQA.
\subsection{Training Parameters}
\label{training_para}
Models were fine-tuned with full-parameter. Training proceeded for a single epoch with a batch size of 64 and a maximum sequence length of 2,048 tokens. We initialized the learning rate at $2\times10^{-5}$ and applied a cosine decay schedule. Optimization used AdamW with bfloat16 (bf16) precision. 

\subsection{Prompts}
\label{prompts}
Prompts for generating reasoning paths for WTQ, TabFact, CRT and SciTab are shown in Figure~\ref{fig:reasoning_path_generation}, \ref{fig:tab_prompt}, \ref{fig:reasoning_path_generation} and \ref{fig:scitab_prompt}, respectively. 
Prompts for generating rationales for a textual PRM are presented in  Figure~\ref{fig:cot_rationale}.
Prompts for generating rationales for \textsc{Multi-Dimensional Analysis} and \textsc{Rephrase, React, Analysis} are shown in Figure~\ref{fig:multi-dimensional} and Figure ~\ref{fig:rephrase}, respectively. 
Refer to Figure~\ref{fig:code_verify} for the prompt for data generation for \textsc{GenPRM.}
Lastly, Figure~\ref{fig:llm-as-a-judge} shows the prompt for collecting data for LLM-as-a-judge.

\begin{table*}[t]
\centering
\small
\begin{tabular}{lcccc}
\hline
\hline
\textbf{Method} & WTQ & TabFact & CRT & SciTab \\
\hline
\hline
Upper Bound &82.52 &97.72 &71.01 &88.39 \\
Pass@1 & 57.65 & 85.24 & 51.09 & 57.10 \\
Majority@8 & 64.86 & 88.52 & 56.45 & 62.58 \\
LLM-as-a-Judge (7B) & 59.06 & 85.51 & 50.13 & 58.00 \\
LLM-as-a-Judge (32B) &67.94 &91.48 &55.49 &64.62 \\
CoT-style PRM (7B) & 62.88 & 88.44 & 51.51 & 60.07 \\
CoT-style PRM (32B) &67.48 &91.86 &55.08 &63.36\\
Multi-dimensional (7B) &62.18 &87.89 &51.94 &59.96 \\
Rephrase-React-Analysis (7B) &62.66 &88.69 &52.74 &59.93 \\
GenPRM (7B) &66.09 &90.20 &53.31 &60.64 \\
GenPRM (32B) & 68.92 & 92.42 & 56.00 & 63.78 \\
\hline
\hline
\end{tabular}
\caption{Main results on in-domain (WTQ, TabFact) and out-of-domain (CRT, SciTab) datasets, measured with the \textit{Best@N} metric (\(N=8\)). PRMs with code execution achieve the best overall performance.}
\label{tab:full_results}
\label{tab:lsc}
\end{table*}
\label{sec:appendix}
\subsection{Dataset Statistics and Licenses}
\label{appendix:data_license}
The statistics of datasets we used are shown in the Table~\ref{tab:test_statistics}.
\begin{table}[t]
    \small
    \setlength{\tabcolsep}{6pt}
    \centering    
    \begin{tabular}{ccc}
    \toprule
    Datasets & \#instances & Domain \\
    \midrule
    WTQ & 4344& Wikipedia \\
    TabFact & 2024 & Wikipedia\\
    SCITAB &1224 &  scientific paper \\
     CRT & 728 &Wikipedia  \\
    \bottomrule
    \end{tabular}
       \caption{Test data statistics.}
       \label{tab:test_statistics}
\end{table}
WTQ \cite{pasupat-liang-2015-compositional} is under the license of \textsc{CC-BY-SA-4.0}\footnote{\url{https://creativecommons.org/licenses/by-sa/4.0/}}. TabFact \cite{chen2020tabfactlargescaledatasettablebased}, CRT \cite{zhang-etal-2023-crt} and SCITAB \cite{lu-etal-2023-scitab} are under the \textsc{MIT}\footnote{\url{https://opensource.org/license/mit}} license.
\subsection{Time Efficiency.}
\label{efficiency}

\begin{table}[t]
\centering
\resizebox{0.48\textwidth}{!}{%
\begin{tabular}{lccccc}
\hline
 \textbf{}& \textbf{WTQ} &\textbf{TabFact} &\textbf{CRT} &\textbf{SciTab}\\
\hline
Textual PRM &61.56 &80.60 &56.82 &90.32\\
Multi-Dimensional &82.75 &88.86 &83.32 &112.64\\
Rephrase-React-Analysis &150.78 &133.81 &208.07 &163.73\\
GenPRM w/o code (7B) &107.22&122.41&154.84&188.31 \\
GenPRM (7B) &222.93 & 228.38 & 290.30 &469.45 \\
GenPRM w/o code (32B) &185.96 &159.95 &255.65 &191.29\\
GenPRM (32B) &508.66 & 433.21 &608.91 &529.66 \\
\hline
\end{tabular}}
\caption{Time efficiency (seconds; lower is better) of PRM variants across four tasks. GenPRMs with code have longer inference times than Textual PRMs, MDA, RRA, and GenPRMs w/o code, indicating that code generation substantially increases latency.}
\label{tab:time_efficiency}
\end{table}
Despite dataset-specific variation, PRMs without code (Textual, MDA, RRA, and GenPRM without execution) are consistently cheaper than GenPRM. Code execution introduces significant runtime overhead, especially with larger models. This suggests that integrating code improves performance at the cost of higher latency and longer inference time.

\subsection{Variance in Reasoning Complexity Across Datasets}
\label{appendix:variance_dataset}
Table~\ref{tab:RQ3} reveals a clear in-/out-of-domain split in path structure. WTQ and TabFact predominantly exhibit \emph{consistent} trajectories (86\% and 81\%), i.e., monotone label sequences with at most one change-point. By contrast, CRT and SciTab show far more \emph{inconsistent} paths (37\% and 50\%), with multiple flips between correct and incorrect steps. This suggests fewer spurious/noisy steps in the in-domain sets (WTQ/TabFact) and a higher noise ratio in CRT/SciTab.

\subsection{Examples of Inflated Scores and Over-Penalization}
\label{case_study}
Examples with inflated and penalized scores are shown in  Figure~\ref{fig:inflated} and Figure~\ref{fig:over_penalized}, respectively.


\label{sec:RQ3}
\begin{table}[t]
\centering
\small
\begin{tabular}{lcccc}
\hline
\textbf{Reasoning Paths} & WTQ &TabFact &CRT &SciTab\\
\hline
Total &100 &100 &100 &100\\
Consistent & 86 &81& 63 &50 \\
Inconsistent & 14 &19 & 37 &50 \\
\hline
\end{tabular}
\caption{Distribution of human-annotated reasoning paths by consistency. A path is \textbf{consistent} if its labels are \textbf{monotonically decreasing} (e.g., $[1,1,1,0,0]$); paths that are always correct or always wrong also count as consistent. A path is \textbf{inconsistent} if it exhibits any $0\!\to\!1$ jump or multiple flips (e.g., $[0,1,1,0,1]$).}

\label{tab:RQ3}
\end{table}

\begin{figure*}[h]
\centering
\begin{wideheaderbox}{Chain with Locally Correct Steps but an Incorrect Final Answer}
Table is: [['Team', 'No', 'Driver', 'Class', 'Rounds'],\\[1mm]
['Josef Kaufmann Racing', '4', 'Robin Frijns', '', 'All'],\\[1mm]
['Josef Kaufmann Racing', '5', 'Hannes van Asseldonk', 'R', 'All'],\\[1mm]
['Josef Kaufmann Racing', '6', 'Petri Suvanto', 'R', 'All'], \\[1mm]
['M\u00fccke Motorsport', '7', 'Maciej Bernacik', 'R', 'All'],\\[1mm]
['M\u00fccke Motorsport', '8', 'Timmy Hansen', '', 'All'], \\[1mm]
['EuroInternational', '11', 'Daniil Kvyat', 'R', 'All'], \\[1mm]
['EuroInternational', '12', 'Carlos Sainz, Jr.', 'R', 'All'],\\[1mm]
['EuroInternational', '14', 'Michael Lewis', '', 'All'], \\[1mm]
['DAMS', '15', 'Javier Taranc\u00f3n', '', 'All'],\\[1mm]
['DAMS', '16', 'Dustin Sofyan', '', '5'], \\[1mm]
['DAMS', '16', 'Luciano Bacheta', '', '7\u20138'],\\[1mm]
['DAMS', '17', 'Fahmi Ilyas', '', '1\u20136'], \\[1mm]
['DAMS', '17', 'Dustin Sofyan', '', '8'], \\[1mm]
['Eifelland Racing', '18', 'Facundo Regalia', '', 'All'], \\[1mm]
['Eifelland Racing', '19', 'C\u00f4me Ledogar', '', 'All'], \\[1mm]
['Eifelland Racing', '20', 'Marc Coleselli', 'R', 'All'], \\[1mm]
['Fortec Motorsport', '24', 'Jack Harvey', '', 'All'], \\[1mm]
['Fortec Motorsport', '25', 'George Katsinis', '', 'All'], \\[1mm]
['Fortec Motorsport', '26', 'Christof von Gr\u00fcnigen', '', 'All']]\\[1mm]
Question is: how many teams had at least three drivers each for the 2010 formula bmw europe season? \\[1mm]
Step 1: First, we need to identify which teams have at least three drivers each. We'll count the drivers for each team. \textcolor{red}{(\ding{51})}\\[1mm]
Step 2: Josef Kaufmann Racing: 3 drivers (4, 5, 6), \textcolor{red}{(\ding{51})}\\[1mm]
Step 3: M\u00fccke Motorsport: 2 drivers (7, 8), \textcolor{red}{(\ding{51})}\\[1mm]
Step 4: EuroInternational: 3 drivers (11, 12, 14), \textcolor{red}{(\ding{51})}\\[1mm]
Step 5: DAMS: 4 drivers (15, 16, 17, 16), \textcolor{red}{(\ding{51})}\\[1mm]
Step 6: Eifelland Racing: 3 drivers (18, 19, 20), \textcolor{red}{(\ding{51})}\\[1mm]
Step 7: Fortec Motorsport: 3 drivers (24, 25, 26), \textcolor{red}{(\ding{51})}\\[1mm]
Step 8: Count the teams with at least 3 drivers: 4 (Josef Kaufmann Racing, EuroInternational, DAMS, Eifelland Racing, Fortec Motorsport). Final answer: \verb|\\boxed{4}|. \textcolor{red}{(\ding{55})}
\end{wideheaderbox}
\caption{A chain with many locally correct steps but an incorrect final answer, inflating the average step score. Labels here are human-annotated.}
\label{fig:inflated}
\end{figure*}
\begin{figure*}[h]
\centering
\begin{wideheaderbox}{Chain with  Local Errors but Correct Final Answer}
Table is: [['Polling Firm', 'Source', 'Date Published', 'N.Anastasiades', 'G.Lillikas', 'S.Malas', 'Others'],\\[1mm]
['RAI Consultants', '[1][dead link]', '16 September 2012', '37.2\%', '14.2\%', '21.9\%', '1.5\%'],\\[1mm]
['Evresis', '[2]', '18 September 2012', '35.2\%', '17.5\%', '19.7\%', '1.7\%'],\\[1mm]
['Noverna', '[3]', '23 September 2012', '35.02\%', '15.81\%', '17.78\%', ''],\\[1mm]
['Prime Consulting Ltd', '[4]', '7 October 2012', '34.7\%', '17.4\%', '18.5\%', ''],\\[1mm]
['CMR Cypronetwork / Cybc', '[5][dead link]', '18 October 2012', '36.9\%', '17\%', '23.8\%', '1.2\%'],\\[1mm]
['Evresis', '[6]', '2 November 2012', '36.9\%', '17.7\%', '20.6\%', '1.4\%'],\\[1mm]
['RAI Consultants', '[7]', '4 November 2012', '38.8\%', '19.8\%', '21.1\%', '2.3\%'],\\[1mm]
['CMR Cypronetwork / Cybc', '[8]', '15 November 2012', '36.8\%', '18.9\%', '22.8\%', '1.6\%'],\\[1mm]
['Prime Consulting Ltd', '[9]', '18 November 2012', '35.9\%', '18.7\%', '19.6\%', '0.6\%'],\\[1mm]
['Evresis', '[10]', '27 November 2012', '37.1\%', '19.6\%', '20.8\%', '0.6\%'], \\[1mm]
['Noverna', '[11]', '2 December 2012', '35.6\%', '17.2\%', '18.1\%', '4.1\%'],\\[1mm]
['Prime Consulting Ltd', '[12]', '3 December 2012', '35\%', '19.1\%', '18.6\%', '1.4\%'], \\[1mm]
['CMR Cypronetwork / Cybc', '[13][dead link]', '17 December 2012', '37.1\%', '20.4\%', '23.1\%', '3.1\%'],\\[1mm]
['Evresis', '[14]', '22 December 2012', '37.4\%', '19.8\%', '21.8\%', '0.5\%'], \\[1mm]
['RAI Consultants Ltd', '[15][dead link]', '13 January 2013', '40.3\%', '17.9\%', '20.5\%', '6.1\%'], \\[1mm]
['CMR Cypronetwork / Cybc', '[16]', '17 January 2013', '38\%', '19.7\%', '23.7\%', '2.7\%'], \\[1mm]
['Prime Consulting Ltd', '[17]', '27 January 2013', '39.2\%', '18.8\%', '19.8\%', '4\%'], \\[1mm]
['Evresis', '[18]', '1 February 2013', '40.8\%', '19.9\%', '22.2\%', '2.5\%'], \\[1mm]
['Prime Consulting Ltd', '[19]', '4 February 2013', '39.8\%', '19.3\%', '20\%', '3\%'],\\[1mm]
['Prime Consulting Ltd', '[20]', '9 February 2013', '40.6\%', '19.6\%', '20.4\%', '2.9\%'], \\[1mm]
['RAI Consultants Ltd', '[21]', '9 February 2013', '42.1\%', '19.4\%', '21.1\%', '4.4\%'], \\[1mm]
['CMR Cypronetwork / Cybc', '[22]', '9 February 2013', '39.9\%', '20.2\%', '24.2\%', '3\%'], \\[1mm]
['Average (only valid votes)', '\u2013', '\u2013', '48.4\%', '22.52\%', '25.29\%', '3.79\%']]\\[1mm]
Question is: which poll did anastasiades win by a larger margin, feb 1 2013 or dec 3 2012?\\[1mm]
Step 1: Identify the relevant polls for comparison - February 1, 2013 poll by Prime Consulting Ltd - December 3, 2012 poll by Prime Consulting Ltd, \textcolor{red}{(\ding{55})}\\[1mm]
Step 2: Extract the percentages for N. Anastasiades from these two polls - February 1, 2013: 40.6\% - December 3, 2012: 35.6\%, \textcolor{red}{(\ding{55})}\\[1mm]
Step 3: Calculate the difference between N. Anastasiades and the second-place candidate (G. Lillikas) for each poll- For February 1, 2013: 40.6\% - 19.6\% = 21.0\% - For December 3, 2012: 35.6\% - 17.2\% = 18.4\%, \textcolor{red}{(\ding{55})}\\[1mm]
Step 4: Compare the margins - February 1, 2013: 21.0\% - December 3, 2012: 18.4\%, \textcolor{red}{(\ding{55})}\\[1mm]
Step 5: Conclude which poll had the larger margin - The February 1, 2013 poll shows a larger margin for N. Anastasiades. Final answer: \verb|\\boxed{February 1, 2013}|
], \textcolor{red}{(\ding{51})}\\[1mm]
\end{wideheaderbox}
\caption{An example of a chain with local errors that does not affect the final answer; in this case, the chain’s score is over-penalized. Labels here are human-annotated.}
\label{fig:over_penalized}
\end{figure*}

\begin{figure*}[h]
\centering
\begin{wideheaderbox}{Reasoning Path Generation prompt on WTQ and CRT}
Please inspect the table(s) and then provide an Answer to the question.  Please reason step by step. Attention: You MUST put your final answer using the following format: Final answer: \verb|\\boxed{final answer}|. Attention: You MUST put your final answer using the following format: Final answer: \verb|\\boxed{final answer}|.
\end{wideheaderbox}
\caption{Prompt for Generating Reasoning Paths on WTQ and CRT Dataset.}
\label{fig:reasoning_path_generation}
\end{figure*}
\begin{figure*}[h]
\centering
\begin{wideheaderbox}{Reasoning Path Generation prompt on TabFact}
Please inspect the table(s) and then provide a True or False answer to the question.  Please reason step by step. Attention: You MUST put your final answer using the following format: Final answer: \verb|\\boxed{True/False}|, Attention: You MUST put your final answer using the following format: Final answer: \verb|\\boxed{True/False}|.
\end{wideheaderbox}
\caption{Prompt for Generating Reasoning Paths on TabFact Dataset.}
\label{fig:tab_prompt}
\end{figure*}
\begin{figure*}[h]
\centering
\begin{wideheaderbox}{Reasoning Path Generation prompt on SciTab}
Please inspect the table(s) and then provide an Answer to the question.  Please reason step by step. Attention: You MUST put your final answer using the following format: Final answer: \verb|\\boxed{True/False/not enough info}|, Attention: You MUST put your final answer using the following format: Final answer:\verb|\\boxed{True/False/not enough info}|.
\end{wideheaderbox}
\caption{Prompt for Generating Reasoning Paths on SciTab Dataset.}
\label{fig:scitab_prompt}
\end{figure*}
\newpage
\begin{figure*}[h]
\centering
\begin{wideheaderbox}{Prompt for generating CoT analysis}
The following is the table question answering problem and a solution (split into paragraphs, enclosed with tags and indexed from 1):\\[1mm]
[Table Question Answering Problem]\\[1mm]
\{problem\}\\[1mm]
[Solution]\\[1mm]
\{solution section\}\\[1mm]
Your task is to verify the correctness of the paragraph in the solution.  Split your verification by `\#\#\# Paragraph \{ID\}`.\\[1mm]
Your verification for each paragraph should be constructed by 2 parts, wrapped by '<analyze></analyze>' and '<output></output>' separately.\\[1mm]
1. In `<analyze>` part, you need to analyze the reasoning process and explain why the paragraph is correct or incorrect in detail.\\[1mm]
2. In `<output>` part, judge if this paragraph is correct or incorrect and put the answer into `<output>` part, e.g., ``<output> **Judgement**: \verb|\\boxed{Yes/No}| </output>``. Every Paragraph must have an `<output>` part.
\end{wideheaderbox}
\caption{Prompt for generating COT-style rationales.}
\label{fig:cot_rationale}
\end{figure*}
\newpage
\begin{figure*}[h]
\centering
\begin{wideheaderbox}{Prompt for generating Multi-Dimensional Rationales}
The following is the table question answering problem and a solution (split into paragraphs, enclosed with tags and indexed from 1):\\[1mm]
[Table Question Answering Problem]\\[1mm]
\{problem\}\\[1mm]
[Solution]\\[1mm]
\{solution Section\}\\[1mm]
Your task is to verify the logical and factual correctness of each solution paragraph.\\[1mm]
Split your verification by '\#\#\#Paragraph {{ID}}'.\\[1mm]
Your verification for each paragraph should be constructed by 2 parts, wrapped by '<analyze></analyze>',  and '<output></output>' separately.\\[1mm]
1. In `<analyze>` part, address these five aspects:\\[1mm]
Part 1: **Restatement**: Verbalize the claim, statement, or answer this paragraph makes.\\[1mm]
Part 2. **Data Check**: Traverse the table and quote row and column to show where the evidence comes from.\\[1mm]
Part 3.  **Logical Consistency**: Check if the claim is logically valid.\\[1mm]
Part 4.  **Numeric Accuracy**: Check if calculations such as sums, percentages, and table lookups, are computed correctly.\\[1mm]
Part 5. **Format**: Verify that the final paragraph puts the final answer using \verb|\\boxed{Final Answer}|.\\[1mm]
If any aspect is wrong, then the paragraph is wrong.\\[1mm]
2. In `<output>` part, judge if this paragraph is correct or incorrect and put the answer into `<output>` part, e.g., ''<output>**Judgement**: \verb|$\\boxed{Yes/No}|</output>''. Every Paragraph must have an `<output>` part.
\end{wideheaderbox}
\caption{Prompt used for generating analyses from five distinct perspective.}
\label{fig:multi-dimensional}
\end{figure*}
\begin{figure*}[h]
\centering
\begin{wideheaderbox}{Prompt for generating Rephrase, React, and Analysis}
The following is the table question answering problem and a solution (split into paragraphs, enclosed with tags and indexed from 1):\\[1mm][Table Question Answering Problem]\\[1mm]\{problem\}\\[1mm][Solution]\\[1mm]\{solution section\}\\[1mm]Your task is to verify the correctness of paragraphs in the solution. Split your verification by `\#\#\# Paragraph \{ID\}`. \\[1mm]Your verification for each paragraph should be constructed by 4 parts, wrapped by `<rephrase></rephrase>`, `<react></react>`, `<analyze></analyze>` and `<output></output>` separately.\\[1mm]1. In `<rephrase>` part,  rephrase the paragraph as a clear, self-contained question, preserving all information from the original paragraph.\\[1mm]2. In `<react>` part, use alternating steps of **Thought**, **Action**, **Observation** to systematically solve the problem from <rephrase> part.\\[1mm]**Thought**: Based on the information currently available, reason through the problem and determine the goal of the next action.\\[1mm]**Action**: Each action must be one of the following five types: 1. LOOKUP\_ROW[row]: Find a row in the table whose label matches the given text. 2. LOOKUP\_COLUMN[column]: Find a column in the table whose header matches the given text. 3. READ\_CELL[row, column]: Retrieve the value in the specified row and column. 4. COMPUTATION[values]: Perform arithmetic or other clearly defined mathematical operations on the provided values. 5. Finish[answer]: Once a clear answer has been determined, use this action to return the answer and terminate the task.\\[1mm]**Observation**: Record the factual result of the action.\\[1mm]3. In `<analyze>` part, based on '<rephrase>' and '<react>' results, analyze in detail if the current paragraph is correct or incorrect, citing the relevant evidence.\\[1mm]4. In `<output>` part, make a final judgement and put the judgement into `<output>` part, e.g., ``<output> **Judgement**: \verb|$\\boxed{Yes/No}| </output>``.
\end{wideheaderbox}
\newpage
\caption{Prompt for generating Rephrase, React, and Analysis rationales.}
\label{fig:rephrase}
\end{figure*}
\begin{figure*}[h]
\centering
\begin{wideheaderbox}{Prompt for generating analysis with code}
The following is the table question answering problem and a solution (split into paragraphs, enclosed with tags and indexed from 1):\\[1mm]
[Table Question Answering Problem]\\[1mm]
\{problem\}\\[1mm]
[Solution]\\[1mm]
\{solution section\}\\[1mm]
Your task is to verify the correctness of paragraph in the solution.  Split your verification by `\#\#\# Paragraph \{ID\}`.\\[1mm]
Your verification for each paragraph should be constructed by 3 parts, wrapped by '<analyze></analyze>', '<verify></verify>' and '<output></output>' separately.\\[1mm]
1. In `<analyze>` part, you need to analyze the reasoning process and explain why the paragraph is correct or incorrect in detail.\\[1mm]
2. In '<verify>' part, you must write **Python code** in the form of ```python \{CODE\}``` to verify every details in the current paragraph that can be verified by code. You must use python code to verify every details in the paragraph. Make sure to print the critic results in the code. Every code will be executed automatically by system. You need to analyze the `[Code Output]` after code executing. Pay attention that you must follow the format of ```python\{CODE\}``` when you write the code, otherwise the code will not be executed.\\[1mm]
3. In `<output>` part, judge if this paragraph is correct or incorrect and put the answer into `<output>` part, e.g., ``<output> **Judgement**: \verb|\\boxed{Yes/No}| </output>``. Every Paragraph must have an `<output>` part.
\end{wideheaderbox}
\caption{Prompt for generating analysis with code execution.}
\label{fig:code_verify}
\end{figure*}
\begin{figure*}[h]
\centering
\begin{wideheaderbox}{Prompt for LLM-as-a-Judge}
The following is the table question answering problem and a solution (split into paragraphs, enclosed with tags and indexed from 1):\\[1mm]
[Table Question Answering Problem]\\[1mm]
\{problem\}\\[1mm]
[Solution]\\[1mm]
\{solution section\}\\[1mm]
Your task is to verify the correctness of paragraph in the solution.  Split your verification by `\#\#\# Paragraph \{ID\}`.\\[1mm]
Your verification for each paragraph should be constructed by 2 parts, wrapped by `<analyze></analyze>` and `<output></output>` separately.\\[1mm]
1. In `<analyze>` part, you need to analyze the reasoning process and explain why the paragraph is correct or incorrect in detail.\\[1mm]
2. In `<output>` part, judge if this paragraph is correct or incorrect and put the answer into `<output>` part, e.g., ``<output> **Judgement**: \verb|$\\boxed{Yes/No}| </output>``. Every Paragraph must have an `<output>` part.'
\end{wideheaderbox}
\caption{Prompt for step-wise LLM-as-a-Judge.}
\label{fig:llm-as-a-judge}
\end{figure*}
\end{document}